\documentclass[10pt,journal,compsoc]{IEEEtran}
\usepackage{luca_style}

\title{AriEL: volume coding for sentence generation}

\author{Luca Celotti*\thanks{* These authors contributed equally.}\ , \hspace{1ex} Simon Brodeur* \ , \hspace{1ex} Jean Rouat\\
NECOTIS, Département GEGI, 
Université de Sherbrooke, 
Québec, Canada \\
\texttt{ \{ luca.celotti, simon.brodeur, jean.rouat \}@usherbrooke.ca} \\
}

\begin{document}

\maketitle

\begin{abstract}
Mapping sequences of discrete data to a point in a continuous space makes it difficult to retrieve those sequences via random sampling. 
Mapping the input to a volume would make it easier to retrieve at test time, and that's the strategy followed by the family of approaches based on Variational Autoencoder. However the fact that they are at the same time optimizing for prediction and for smoothness of representation, forces them to trade-off between the two.
We improve on the performance of some of the standard methods in deep learning to generate sentences by uniformly sampling a continuous space.
We do it by proposing AriEL, that constructs volumes in a continuous space, without the need of encouraging the creation of volumes through the loss function. We first benchmark on a toy grammar, that allows to automatically evaluate the language learned and generated by the models. Then, we benchmark on a real dataset of human dialogues.
Our results indicate that the random access to the stored information is dramatically improved, and our method AriEL is able to generate a wider variety of correct language by randomly sampling the latent space.
VAE follows in performance for the toy dataset while, AE and Transformer follow for the real dataset. This partially supports to the hypothesis that encoding information into volumes instead of into points, can lead to improved retrieval of learned information with random sampling. This can lead to better generators and we also discuss potential disadvantages.
\end{abstract}

\section{Introduction}


It is standard for neural networks to map an input to a point in a $d$-dimensional real space \citep{hochreiter1997long, vaswani2017attention, lecun1989backpropagation}. However, that makes it difficult to find a specific point when the real space is being sampled randomly.
That can limit the applicability of pre-trained models to their initial scope.
Some approaches do map an input into volumes in the latent space.
The family of approaches that stem out of the idea of Variational Autoencoders \citep{Kingma2014, Bowman2016, rezende2015variational, chen2018neural} are trained to encourage such type of representations. By encoding an input into a probability distribution that is sampled before decoding, several neighbouring points in $\mathbb{R}^{d}$ can end up representing the same input.

However, it often implies having two summands in the loss, a Kullback–Leibler divergence term and a log-likelihood term \cite{Kingma2014, Bowman2016}, that fight for two different causes. In fact, if we want a smooth and volumetric representation, encouraged by the KL loss, it might come at the cost of having worse reconstruction or classification, encouraged by the log-likelihood. Therefore, each diminishes the strength and influence of the other.

By giving partially up on the smoothness of the representation, we propose instead a method to explicitly construct volumes, without a loss that is implicitly encouraging such behavior. We propose AriEL, a method to map sentences to volumes in $\mathbb{R}^{d}$ for efficient retrieval with either random sampling, or a network that operates in its continuous space. It draws inspiration from arithmetic coding (AC) and k-d trees (KdT), and we name it after them \textit{Arithmetic coding and k-d trEes for Language} (AriEL). For simplicity we choose to focus on language, even though the technique is applicable for the coding of any variable length sequence of discrete symbols. We prove how such a volume representation eases the retrieval of stored learned patterns.

AC is one of the most efficient lossless data compression techniques \citep{witten1987arithmetic, arithmetic1963}. As illustrated in figure \ref{fig:grammar_arithmetic_coding}, AC assigns a sequence to a segment in [0,1] whose length is proportional to the frequency of that sequence in the dataset.
KdT \citep{bentley1975multidimensional} is a data structure for storage that can handle different types of queries efficiently. It is typically used as a fast approximation to k-nearest neighbours in low dimensions \citep{friedman1977algorithm}. It organizes data in space according to which half of the space it belongs to with respect to the median. Then it moves to the following of the k axis and repeats the process of splitting with respect to the median and turning to a new axis.

Our contributions are therefore:
\begin{itemize}
    \item AriEL, a volume coding technique based on arithmetic coding \citep{arithmetic1963} and k-d trees \citep{bentley1975multidimensional} (Section \ref{sec:arithmetic}), to improve the retrieval of learned patterns with random sampling;
    \item the use of a context-free grammar and a random bias in the dataset (Section \ref{sec:dataset}), that allow us to automatically quantify the quality of the generated language;
    \item the notion that explicit volume coding (Section \ref{sec:relatedwork} and \ref{sec:discussion}) can be a useful technique in tasks that involve the generation of sequences of discrete symbols, such as sentences.
\end{itemize}

\section{Related Work}
\label{sec:relatedwork}

\noindent\textbf{Volume codes:}
We define a \textit{volume code} as a pair of functions, an encoding and a decoding functions, where the encoding function maps an input $x$ into a set that contains compact and connected sets of $\mathbb{R}^{d}$ \citep{munkres2018elements}, and the decoding function maps every point within that set back to $x$. It is a form of distributed representations \cite{hinton1984distributed} in the sense that the latter only assumes that the input $x$ will be represented as a point in $\mathbb{R}^{d}$. For simplicity, we define \textit{point codes} as its complementary, the distributed representations that are not volume codes, and therefore map $x$ to isolated points in $\mathbb{R}^{d}$.
Volume codes differ from coarse coding \cite{hinton1984distributed} in the sense that in this case the code is represented by a list of zeros and ones that identifies in which overlapping sets the $x$ falls into.
Both generative and discriminative models
\cite{ng2002discriminative, Kingma2014, jebara2012machine} can end up learning volume codes by encouraging smoothness of representation via the loss function \citep{bengio2013representation}. We call \textit{implicit volume code}, the latter, when the volume code is encouraged through the loss function. We call \textit{explicit volume code}, when the volumes are constructed instead through the arrangement of neurons in the network.


\noindent\textbf{Sentence} generation through random sampling:
Generative Adversarial Networks (GAN) \cite{Goodfellow2014} are generative models conceived to map random samples to a learned generation through a 2-players game training procedure. They have had in the past trouble for text generation, due to the non differentiability of the $argmax$ performed at the end of the generator, and partially generated sequences are non trivial to score \cite{yu2017seqgan}. Several advances have significantly improved the performance of this technique for text generation, such as using the generator as a reinforcement learning agent trained on next symbol generation through Policy Gradient \cite{yu2017seqgan}, avoiding a binary classification typical in GAN in favor of a cross-entropy for the discriminator that evaluates each word generated \cite{xu2018diversity}, or with the Gumbel-Softmax distribution \cite{kusner2016gans}. 
Random sampling the latent space is used as well by Variational Autoencoders (VAE) \cite{Kingma2014}, to smooth the representation of the learned patterns. Training VAE for text has been shown to be possible with KL annealing and word dropout \cite{Bowman2016}, and made easier with convolutional decoders \cite{46890, yang2017improved}. An important line of research has focused on generalizing VAE to more flexible priors, through techniques such as Normalizing Flows \cite{rezende2015variational} or Inverse Autoregressive Flows \cite{kingma2016improved}. Several works explore how VAE and GAN can be combined \cite{makhzani2015adversarial, tolstikhin2017wasserstein, mescheder2017adversarial} where GAN provides VAE with a learnable prior distribution, and VAE provides GAN with a more stable training procedure.
We define AriEL to be used in combination with the previously mentioned methods, since it can be used as a generator or a discriminator in a GAN, or as an encoder or a decoder in an autoencoder. However it differs from them in the explicit procedure to construct volumes in the latent space that correspond to different inputs. The intention is to fill the entire latent space with the learned patterns, to make them easy to retrieve by uniform random sampling.

\noindent\textbf{Arithmetic coding and neural networks:}
AC has been used for neural network compression in \cite{wiedemann2019deepcabac} but typically, neural networks are used in AC as the model of the data distribution, to perform prediction based compression,
for real time speech compressed transmission \cite{pasero2003neural}, image compression \cite{triantafyllidis2002neural, jiang1993lossless}, high efficiency video coding \cite{ma2019neural} and for general purpose compression \cite{tatwawadi2018deepzip}.
We turn AC into a compression algorithm in $d$ real numbers, to combine its properties with the properties of high-dimensional spaces, which is the domain of neural networks.

\noindent\textbf{K-d trees and neural networks:}
Neural networks are typically used in conjuction with KdT to reduce the dimensionality of the search space, for KdT to be able to perform queries efficiently \cite{woodbridge2018detecting, yin2017efficient, vasudevan2009gaussian}. KdT have been substituted completely by neural networks in \cite{cheng2018deep} to make better use of the limited memory resources in low-performance hardware. KdT has been used as well in combination with Delaunay triangulation for function learning, as an alternative to NN with Backpropagation \cite{gross1995kd}. A KdT inspired algorithm is used in \cite{maillard1994neural} to guide the creation of neurons to grow a neural network.
We use KdT to make sure that when we turn AC into a multidimensional version of itself, it splits in a systematic way all dimensions of $\mathbb{R}^{d}$, so it can make use of all the space available.

\section{Methodology}

\begin{figure}[h!]
\centering
\includegraphics[width=.95\linewidth]{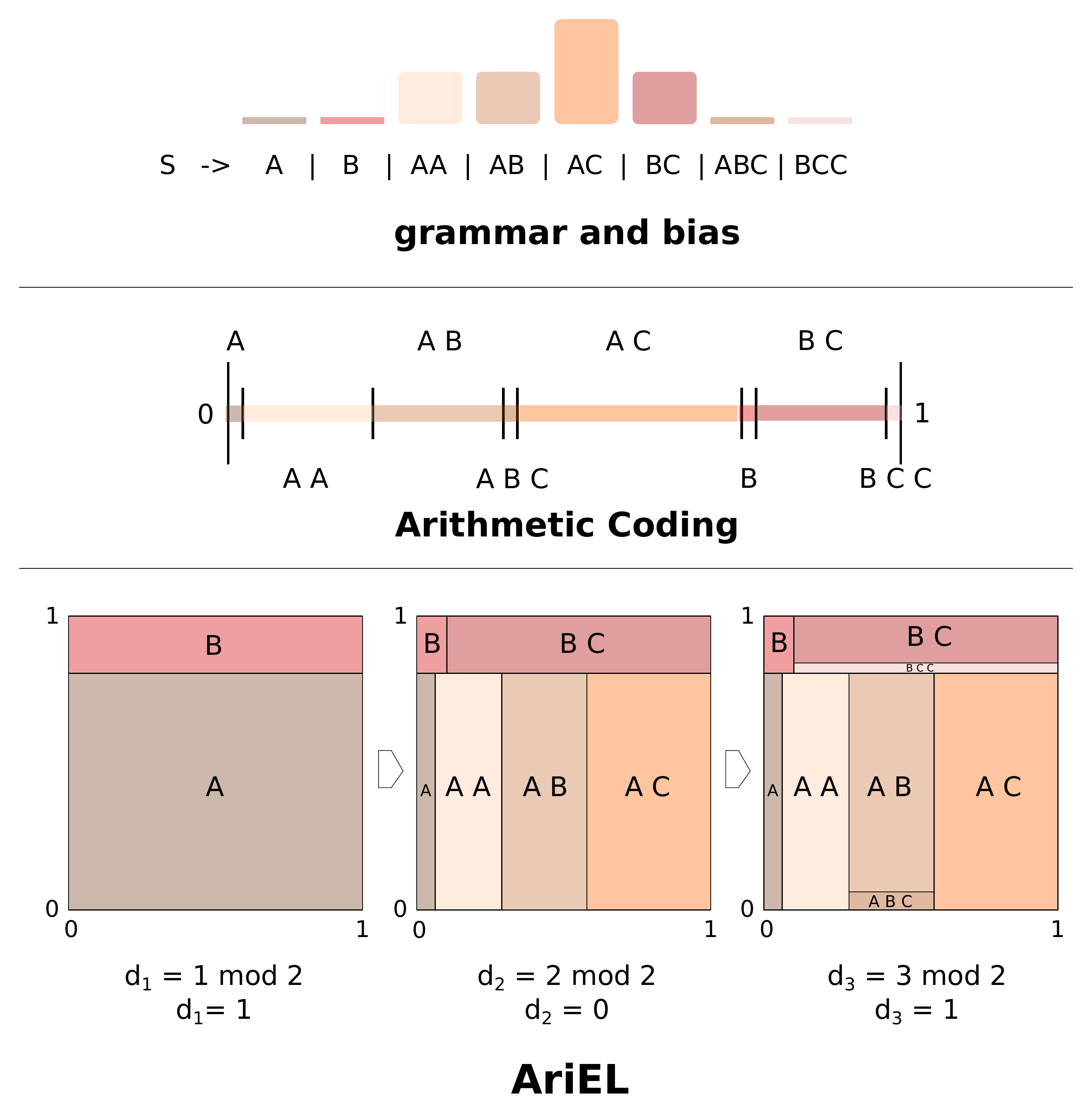}
\vspace{-.3cm}
\cprotect\caption{\textbf{Sentence embedding with arithmetic coding and AriEL.} In this example, the generating context-free grammar (CFG) is $S \rightarrow A | B | A A | A B | A C | B C | A B C | B C C$, and the bar plot on top indicates the frequency of those sentences in the dataset, as an extra bias to the language. Standard arithmetic coding (in the middle) encodes any sequence of this CFG over a single dimension in the interval $[0, 1]$, and the frequency of the sentence determines the length of the range assigned on that segment. AriEL is designed as a multidimensional extension of AC (illustrated in 2D), where the frequency information is preserved in the volumes. As the symbols of the sentences are read by the Language Model, this provides the boundaries where the next symbols are to be found. For a 2-dimensional latent space, $d=2$, the axis to split to find symbol $s_i$ is $d_i = i \; \text{mod} \; d$. In the image $d_i=0$ represents the horizontal axis, while $d_i=1$ represents the vertical axis.}
\label{fig:grammar_arithmetic_coding}
\end{figure}

\begin{algorithm}
\caption{\textbf{AriEL Encoding}  \\ From sentence to continuous space. B stands for bound, and $B_{up}$ and $B_{low}$ for the upper and lower bounds that define the AriEL volumes, the blue color identifies the lines with the major differences between encoder and decoder and $P_{LM}$ identifies the Language Model inside AriEL. Its cumulative distributions ($c_{up}$, $c_{low}$) are used to define the limits of the volumes and its size ($range$). Finally the volumes are represented by their central point $\textbf{z}$ for simplicity.\vspace{-.32cm}
\label{alg:encoding}}
\begin{algorithmic}[1]
\Require{ sentence: $S=(s_j)_{j=1}^n$}
\Ensure{$\textbf{z}$ represents $S$ in $[0,1] ^{d}$}
\Statex
\Procedure{AriEL\_encode}{$ S$}    
\State $d = $ latent space dimension
\State $B_{low} = zeros(d)$
\State $B_{up} = ones(d)$
\State $n = length(S)$
\Statex
\For{$i = 0, \cdots, n-1$}
    \Statex \Comment{\textbf{choose the dimension to split}}
    \State{$d_i = i \mod d$}
    \State $P_{next}(s) = \textcolor{NiceOrange}{P_{LM}}(s|(s_j)_{j<i})$
    \State $c_{low}(s)=\sum_{s>s'} P_{next}(s')$
    \State $c_{up}(s)=\sum_{s>s'-1} P_{next}(s')$
    \State $range=B_{up}(d_i) - B_{low}(d_i)$
    \Statex \Comment{\textbf{update the bounds of the volume}}
    \State $B_{up}(d_i)=B_{low}(d_i)+range\cdot c_{up}(s_i)$ 
    \State $B_{low}(d_i)=B_{low}(d_i)+range\cdot c_{low}(s_i)$
\EndFor
\Statex \Comment{\textbf{represent the volume with its central point}}
\State $\textcolor{NiceBlue}{\textbf{z} = (B_{low} + B_{up})/2}$ 
\State\Return $\textbf{z}$
\EndProcedure
\end{algorithmic}
\end{algorithm}

\subsection{AriEL: volume coding of language in continuous spaces}
\label{sec:arithmetic}


\begin{algorithm}
\caption{\textbf{AriEL Decoding}  \\ From continuous space to sentence. B stands for bound, and $B_{up}$ and $B_{low}$ for the upper and lower bounds that define the AriEL volumes, the blue color identifies the lines with the major differences between encoder and decoder and $P_{LM}$ identifies the Language Model inside AriEL. Its cumulative distributions ($c_{up}$, $c_{low}$) are used to define the limits of the volumes and its size ($range$). \textbf{z} is used to identify which volume has to be picked next. \label{alg:decoding}}
\begin{algorithmic}[1]
\Require{$\textbf{z}$ represents $S$ in $[0,1] ^{d}$}
\Ensure{ sentence: $S=(s_j)_{j=1}^n$}
\Statex
\Procedure{AriEL\_decode}{$ \textbf{z} $}   
\State $d = dimension( \textbf{z} )$
\State $B_{low} = zeros(d)$
\State $B_{up} = ones(d)$
\Statex
\State $\textcolor{NiceBlue}{S = \langle START \rangle}$
\For{$i = 0, \cdots, \textcolor{NiceBlue}{n_{max}-1}$}
    \Statex \Comment{\textbf{choose the dimension to unsplit}}
    \State{$d_i = i \mod d$}
    \State $P_{next}(s) = \textcolor{NiceOrange}{P_{LM}}(s|\textcolor{NiceBlue}{S})$
    \State $c_{low}(s)=\sum_{s>s'} P_{next}(s')$
    \State $c_{up}(s)=\sum_{s>s'-1} P_{next}(s')$
    \State $range=B_{up}(d_i)-B_{low}(d_i)$
    \Statex \Comment{\textbf{update the bounds of the volume}}
    \State $\textcolor{NiceBlue}{Bs_{up}(s)}=B_{up}(d_i)+range\cdot c_{up}(\textcolor{NiceBlue}{s})$ 
    \State $\textcolor{NiceBlue}{Bs_{low}(s)}=B_{low}(d_i)+range\cdot c_{low}(\textcolor{NiceBlue}{s})$
    \Statex \Comment{\textbf{any point in the volume is assigned the symbol $s_i$}}
    \State $\textcolor{NiceBlue}{s_i = find_s\Big(Bs_{low}(s)<\textbf{z}(d_i)<Bs_{up}(s)\Big)}$
    \State $B_{up}(d_i) = \textcolor{NiceBlue}{Bs_{up}(s_i)}$
    \State $B_{low}(d_i) = \textcolor{NiceBlue}{Bs_{low}(s_i)}$
    \State $\textcolor{NiceBlue}{S = S.append(s_i)}$
\EndFor
\State\Return $S$
\EndProcedure
\end{algorithmic}
\end{algorithm}

AriEL maps the sentence $(s_1, \cdots, s_n)$ to a $d$-dimensional volume of $P((s_1, \cdots, s_n)) = \Pi_{i=1}^n P(s_i | (s_j)_{j<i})$. The sentence is encoded as the center of that volume for simplicity, and any point within it is decoded to the same sentence. Decoding  iteratively computes the bounds of the volumes for all possible next symbols  and checks inside which bounds the vector is, to find the next symbol at each step. The exact algorithm is described in algorithm \ref{alg:encoding} and \ref{alg:decoding}.

To adapt KdT to more splits than binary, we decide to split the chosen dimension giving a space from 0 to 1 to each possible next symbol, proportional to the probability of that next symbol: the first symbol in the sentence will be assigned a segment of length $P(s_1)$ in the first axis chosen $d_1$, and next symbols will be assigned a segment proportional to their probability conditional to the symbols previously seen e.g. $P(s_3|(s_2,s_1))$ on the axis $d_3$, where $s_1$, $s_2$ and $s_3$ are the first three symbols in the sentence. Then turn to the following axis and continue the process of splitting and turning.

In figure \ref{fig:grammar_arithmetic_coding} for example, the possible symbols in the dataset are $s_i \in \{ A, B, C, \text{EOS}\}$, where EOS stands for End-Of-Sentence.
The initial token $s_1=A$ is given a portion on the axis $d_1$ of length $P(A)$, larger than the portion given to $s_1=B$ or $s_1=C$, since there are less sentences that start with $B$ than with $A$, and there is none that starts with $C$: $P(A)>P(B)>P(C)=0$. Then, we split the axis $d_2$ according to the probability of the next symbol being $s_2 = A, B, C, \text{EOS}$. In this case the second most likely symbol after symbol $s_1=A$ is $s_2=C$, and that is why $AC$ ends with a larger volume than $AA$, $AB$ and $A$ (which is an abbreviation for \lq$A\ \text{EOS}$\rq). As the sentences become longer than the dimensionality $d$ chosen, next symbols will be assigned an axis $d_3$ that has been split already, but only the section of interest will be further split. So, in the figure, the sentence ABC will take a portion of AB equal to $P(C|(AB))$, while \lq$AB\ \text{EOS}$\rq \ will take a portion equal to $P(\text{EOS}|(AB))$.

We select the next axis in $\mathbb{R}^{d}$ to split to be $d_i = i \; \text{mod} \; d$, where $i \in \lbrace 1, 2, \ldots, n \rbrace$ and $n$ is the length of the sequence. If $n$ is larger than the dimension $d$, then the segment in $d_i$ previously selected by the splitting process, will be split again. Since we do not have access to the true statistics of the data $P(s_i|(s_j)_{j<i})$, we applied a neural network to approximate that distribution, the Language Model (LM) of AriEL, $P_{LM}(s_i|(s_j)_{j<i})$.
This will approximate the frequency information that makes AC entropically efficient (figure \ref{fig:grammar_arithmetic_coding}) since after a successful training, $P_{LM}(\cdot)$ will converge to $P(\cdot)$. 
AriEL conserves then the arithmetic coding property of assigning larger volume to frequent sentences. 

AriEL only uses a bounded region of $\mathbb{R}^{d}$, the interval $[0,1]^{d}$, so encoder and decoder map each input to a compact set and from a compact set respectively. Moreover, AriEL encoder assigns sequence $x$ to a hyper-rectangle \citep{hyperrectangle} and AriEL decoder assigns values inside that hyper-rectangle to the sequence $x$. Since hyper-rectangles cannot be divided into two disjoint non-empty closed sets, they are connected \citep{munkres2018elements}. Therefore AriEL is a \textit{volume code}. AriEL is an \textit{explicit volume code} since its LM is trained only on a next word prediction log-likelihood loss, without a regularization term, and the volumes are constructed by arranging the outputs of the softmax neurons into a $d$ dimensional grid. However, even if the fact that sentences that start with the same symbols will remain close to each other, making it a smooth representation \citep{bengio2013representation}, it is imposing a prior that might not be suited for every task.

In this work, AriEL's language model neural network consists of a word embedding, followed by a LSTM unit and a feedforward layer that outputs a softmax distribution over next possible symbols. Then the argmax is not applied directly to the softmax, but the probabilities defined by the latter are used as a deterministic russian roulette with the latent space point as the deterministic pointer that chooses the position in the roulette.

The most notable features of AriEL are that (1) if the language model learned the grammar, \textit{almost all} the continuous space encodes \textit{almost only} grammatically correct sentences, and (2) sequences are mapped to a volume, not to a point, therefore small noise in the continuous space will produce the same sentence. Theoretically this is so because a sequence of symbols that are $\epsilon$ likely together, with $\epsilon$ very small, will be assigned an exponentially small volume of $\epsilon^n$. We support experimentally this claim with the generation studies below.
Another feature inherited from KdT is (3) that \textit{any} sequence is assigned a point in the continuous space, no matter how small: in theory this allows AriEL to be able to perfectly encode and decode any sequence to itself. This is studied experimentally with the generalization studies below.

AriEL with a RNN-based language model has a computational complexity of $O(nD^2)$ for both encoding and decoding, as it can be seen in algorithms \ref{alg:encoding} and \ref{alg:decoding}, where $n$ is the length of the sequence and $D$ is the dimensionality of the RNN hidden state. We use capital $D$ to refer to the length of the longest hidden layer among the recurrent layers in the encoder or in the decoder, while $d$ refers to the length of the last hidden layer, the one that defines the size of the latent space. Since the language model only performs short-term (i.e. next word) modelling, AriEL allows the use of a smaller $D$ and thus has significantly less trainable parameters than other seq2seq models. AriEL has a time complexity of $O(n)$ for both encoding and decoding, which is on par with conventional recurrent networks for seq2seq learning.

\subsection{Neural Networks: models and experimental conditions}

We compare AriEL to some of the classical approaches to map variable length discrete spaces to fixed length continuous spaces. These are the sequence to sequence recurrent autoencoders (AE) \citep{Sutskever2014}, their variational version (VAE) \citep{Bowman2016} and Tranformer  \citep{vaswani2017attention}.
All of them are trained for next word prediction of word $s_i$, when all the previous words are given as input, and they are trained over the biased train set, defined in section \ref{sec:dataset}. We applied teacher forcing \citep{6795228} to the decoder during training. All of them can be split into an encoder that maps the sentences at the input into $\mathbb{R}^{d}$, and a decoder that maps back from $\mathbb{R}^{d}$ into a sentence. For all methods, all the input sentence is fed to the encoder, and subsequently the output of the encoder is used to produce the complete decoded output. We call word vector representation, the models that pass vectors that represent words from the encoder to the decoder, such as Transformer, while we call sentence vector representation the models that pass a vector that represents the whole sentence from the encoder to the decoder, such as AE, VAE and AriEL \citep{kiros2015skip}.

For both AE and VAE, we stack two GRU layers \citep{Cho2014} with 128 units  at both, the encoder and the decoder, to increase their representational capabilities \citep{Pascanu2014}. Other recurrent networks gave similar results \citep{hochreiter1997long, Li2018}.
The last encoder layer has  either $d = 16$ units or $d = 512$ for all methods. 
The output of the decoder was a softmax distribution over the entire vocabulary.  AriEL's Language Model was implemented as an embedding layer of dimension 64, followed by an LSTM of 140 units and a Fully Connected layer with a softmax output to predict the next symbol in the sentence.

We compared AriEL, AE and VAE to Tranformer  \citep{vaswani2017attention}, the state-of-the-art in many S2S problems \citep{vaswani2017attention, dai2019transformer, radford2018improving}. Since it is a fixed-length representation at the word level but it is variable-length at the sentence level, we padded all sentences to the maximum length in the dataset to be able to compare its latent space capacity to the other models.
We split the Transformer into encoder and decoder, to study how they make use of the real latent space, and we use its decoder as a generator sampling randomly the input. The literature tends to use only the decoder of the original model \citep{dai2019transformer, radford2018improving} and there is an active interest in widening the applications and in the clarification of the inner mechanisms \cite{kovaleva-etal-2019-revealing, jawahar-etal-2019-bert}. We treat as the latent dimension of the Transformer its $d_{model}$, that will take a value of 16 or 512. We choose most of the other parameters as in the original work \citep{vaswani2017attention}: the number of parallel attention heads as $n_{head}=8$, the key and value dimension as $d_{key}=64$ and $d_{value}=64$, and a dropout regularization of 0.1, and we only change the stack of identical decoders and encoders to $n_{parallel \ layers}=2$, and the inner dimension of the position-wise feed-forward network to $d_{inner \ layer}=256$ to have a number of parameters similar to the other methods.  

We choose the hyper-parameters of AE, VAE, Transformer and AriEL's LM to keep a number of trainable parameters comparable to each other. For the toy dataset $d = 16$ to about 270K parameters. For $d = 512$ we used the same hyper-parameters as before, and only changed the latent dimension. This implied that each model scaled differently: 120M parameters for AE and VAE, 9M for Transformer and AriEL keeps the same small network, since the size of its latent space can be defined at any time without trainable parameters depending on it. During the training on the GuessWhat?! dataset, the scaling with respect to the other methods was different: we trained the Transformer 16 on the GuessWhat?! dataset with  $n_{parallels \ layers}=20$, to have a number of trainable parameters on the same order of magnitude than the other methods, 2,666K rather than the 588K that it had when $n_{parallel \ layers}=2$, but the performance was much worse, so we decided to present the results for the $n_{parallel \ layers}=2$ Transformer. For $d = 512$ again, all of them scaled differently, as it can be seen in table \ref{tab:dataGW}.

We go through the training data 10 times, in mini-batches of 256 sentences.
We use the Adam \citep{Kingma2015} optimizer with a learning rate of 1e-3 and gradient clipping at 0.5 magnitude. During training, the learning was reduced by a factor of 0.2 if the loss function didn't decrease in the last 5 epochs, but with a minimum learning rate of 1e-5. For all RNN-based embeddings, kernel weights used the Xavier uniform initialization \citep{Glorot2010}, while recurrent weights used random orthogonal matrix initialization \citep{Saxe2014}. All biases are initialized to zero. All embedding layers are initialized with a uniform distribution between [-1, 1]. For Transformer all the matrices in the multihead attention and in the position-wise feedforward module, used the Xavier uniform initialization \citep{Glorot2010}, the beta of the layer normalization uses zeros, and its gamma uses ones for initialization. AE and VAE are trained with a word dropout of 0.25 at the input, and VAE is trained with KL loss annealing that moves the weight of the KL loss from zero to one during the 7th epoch, similarly to the original work \citep{Bowman2016}.

\subsection{Datasets: toy and human sentences}
\label{sec:dataset}

We perform our analysis on two datasets. A toy dataset of sentences generated from a context-free grammar and a realistic dataset of sentences written by humans while playing a cooperative game.

\noindent\textbf{The toy dataset:} we generate questions about objects with a context-free grammar (CFG), fully specified in the Supplementary Materials, section \ref{app:cfg}. To stress the learning methods and understand their limits we choose a CFG with a large vocabulary and numerous grammar rules, rather than smaller but more classic alternatives (e.g. REBER). The intention is as well to focus on dialogue agents and that's the reason why all sentences are framed as questions about objects.

In this work we distinguish between \textit{unbiased} sentences, those that have been simply sampled from the CFG, and \textit{biased} sentences, those that after being sampled from the CFG have been selected according to an additional structural constraint. To do so we generate an adjacency matrix of words that can occur together in the same sentence, and we use that as the filter to bias the sentences. For simplicity the adjacency matrix has been generated randomly. The intention is to emulate the setting were a CFG is constrained by realistic scenes, in which case not all the grammatically correct sentences can be semantically correct: e.g. "Is it the wooden toilet in the kitchen ?" could be grammatically correct in a given CFG, but semantically incorrect given that it does not usually happen in a realistic scene. We use it to detect to which degree each learning method is able to extract the grammar and extract the roles of each word, despite a bias that could make this task harder.

The vocabulary consists of 840 words. The maximal length of the sentences is of 19 symbols and the mean length is of 9.9 symbols. We split the biased dataset into 1M train sentences, 10k test sentences and 512 validation sentences, where no sentence is shared between sets. The validation set is small to speed up training. We created another set of 10k unbiased test sentences with the same CFG, where we only gather sentences that don't follow the adjacency matrix, to make sure that the overlap of this test set is zero with previous ones. We train the learning models on the biased sentences and we use the unbiased to test if they were able to grasp the grammar behind.

\noindent\textbf{The real dataset:} we choose the GuessWhat?! dataset \citep{deVries2016}, a dataset of sentences asked by humans to humans to solve a cooperative game. This dataset features a vocabulary of 10,469 words, an order of magnitude larger than the toy CFG. The maximal length of the sentences is of 57 symbols, and the mean length is of 5.9 symbols.

\subsection{Evaluation Metrics} 

We perform a qualitative and a quantitative assessment of the models.

\subsubsection{Qualitative evaluations}

The four qualitative studies are: (1) we list a few samples of reconstruction via next word prediction of unbiased sentences, to understand the generalization capabilities of the different models (table \ref{tab:generalization}), (2) we list a few samples of generated sentences when the latent space is sampled randomly, to understand the generation capabilities (table \ref{tab:generation}), (3) we visualize all the points randomly sampled in the latent space for generation, and we color code them according to the number of adjectives present in the sentence produced if it was grammatically correct, and in another color if it was not grammatically correct, (first row, figure \ref{fig:visualization45}), and (4) we visualize where biased test sentences belonging to different grammar rules and sentence length were mapped by the encoder (second and third row, figure \ref{fig:visualization45}). All qualitative studies are performed for $d=16$.

\subsubsection{Quantitative evaluations on the toy grammar, CFG}

We propose measures that cover 3 properties of an autoencoder: the quality of generation, prediction and generalization. 
We perform our studies for networks with a latent dimension of 16 units, to understand their compression limits, and for a latent dimension of 512 units, which is often taken as the default size in the literature \citep{Kingma2014, vaswani2017attention}.

\noindent\textbf{Generation/Decoding Quality}
is evaluated with sentences produced by the decoder when the latent space of each model is sampled randomly.
The sampling is done uniformly in the continuous latent space, within the maximal hyper-cube defined by the encoded test sentences. 
We sample 10k sentences and apply four measures:
\emph{i)} \textit{grammar coverage (GC)} as the number of grammar rules (e.g. single adjective, multiple adjectives) that could be parsed in the sampled sentences, over four, the maximal number of adjectives plus one for sentences without adjectives;
\emph{ii)} \textit{vocabulary coverage (VC)} as the ratio between the number of words in the vocabulary that appeared in the sampled sentences, over 840, the size of the complete vocabulary;
\emph{iii)} \textit{uniqueness (U)} as a ratio of unique sampled sentences;
and \emph{iv)} \textit{validity (V)} as a ratio of valid sampled sentences, sentences that were unique and grammatically correct. 

\noindent\textbf{Prediction Quality}
is evaluated by encoding the 10k \emph{biased} test sentences and looking at the reconstructions produced by the decoder, using the following objective criteria: \emph{i)} \textit{prediction accuracy biased (PAB)} as a ratio of correctly reconstructed sentences (i.e. all words must match); \emph{ii)} \textit{grammar accuracy (GA)} as a ratio of grammatically correct reconstructions (i.e. can be parsed by the CFG, even if the reconstruction is not accurate). 
and \emph{iii)} \textit{bias Accuracy (BA)} as the ratio of inaccurate reconstructions that are still grammatical and keep the bias of the training set.

\noindent\textbf{Generalization Quality}
is evaluated using the 10k \emph{unbiased} test sentences while the embeddings were trained on the \emph{biased} training set. The \textit{prediction accuracy unbiased (PAU)} is computed in the same way as \textit{PAB}, as the ratio of correctly reconstructed ubiased sentences. It allows us to measure how well the latent space can generalize to grammatically correct sentences outside the language bias.

\subsubsection{Quantitative evaluations on the real dataset, GuessWhat?!}

In a real dataset we don't have a notion of what is grammatically correct, since humans can use spontaneously ungrammatical constructions. We quantified the quality of the language learned with two measures \textit{uniqueness} is the percentage of the sentences generated with random sampling that was unique over the 10K generations and \textit{validity} was the percentage of the unique sentences that could be found in the training data, indicating how easy it was to retrieve the learned information.

\subsubsection{Quantitative evaluations: random interpolations within AriEL}

In figure \ref{fig:imterpolations} we show what we call the interpolation diversity given  the  dimension  of the  latent  space.  It  measures  how  many  of  the  sentences generated  through  a  straight  line  between  two  random points  in $\mathbb{R}^{d}$ were  unique  and  grammatically  correct  for AriEL for  different  values  of d. The Language Model is the one trained on the toy grammar.

\section{Results}

\subsection{Qualitative Evaluations}
\label{sec:qualitative}

We present the qualitative studies performed for $d=16$. Table \ref{tab:generalization} shows the  output of the generalization study. To avoid cherry picking, we display the first 4 reconstructed sentences. AE and VAE fail to generalize to the unbiased language, however both manage to keep the structure at the output of the input sentence, to a large extent. Their behavior improved significantly when the latent space dimension is increased to $d=512$, figure \ref{fig:radars}, with the corresponding increase of parameters. In theory, AriEL is able to reconstruct any sequence by design, by keeping a volume for each of them. However in practice, it failed as well even if slightly less often than the Transformer. Both produce reconstructions of the unbiased input at a similar rate, as it can be see in table \ref{tab:generalization} and in the metric PAU in table \ref{tab:data} and figure \ref{fig:radars}. This means that to a reasonable degree, the areas that represent unseen data during training, are available and relatively easy to track for AriEL and Transformer. Instead, all the latent space seems to be taken almost exclusively by the content of the training set for AE and VAE, since sentences that are not seen during training (in this case the unbiased sentences) cannot be reconstructed at all.

The generation study is shown in Table \ref{tab:generation} (first 4 samples for each model). AriEL excels at this task, and almost all generations are unique and grammatically correct (\textit{valid} in our definition). AE and VAE perform remarkably well given the small latent space. As it will be shown in the quantitative study, VAE almost triples AE performance in terms of generation of valid sentences when $d=16$, validity in table \ref{tab:data}.  Transformer performs poorly at this task, and it is very hard to get grammatical sentences when the latent space is sampled randomly. The quantitative analysis reveals however that with the increase of the latent space, Transformer, AE and VAE achieve all improved validity, remaining at a performance of one third the performance of AriEL.

\begin{figure}[!htb]
\centering
\includegraphics[width=1.\linewidth]{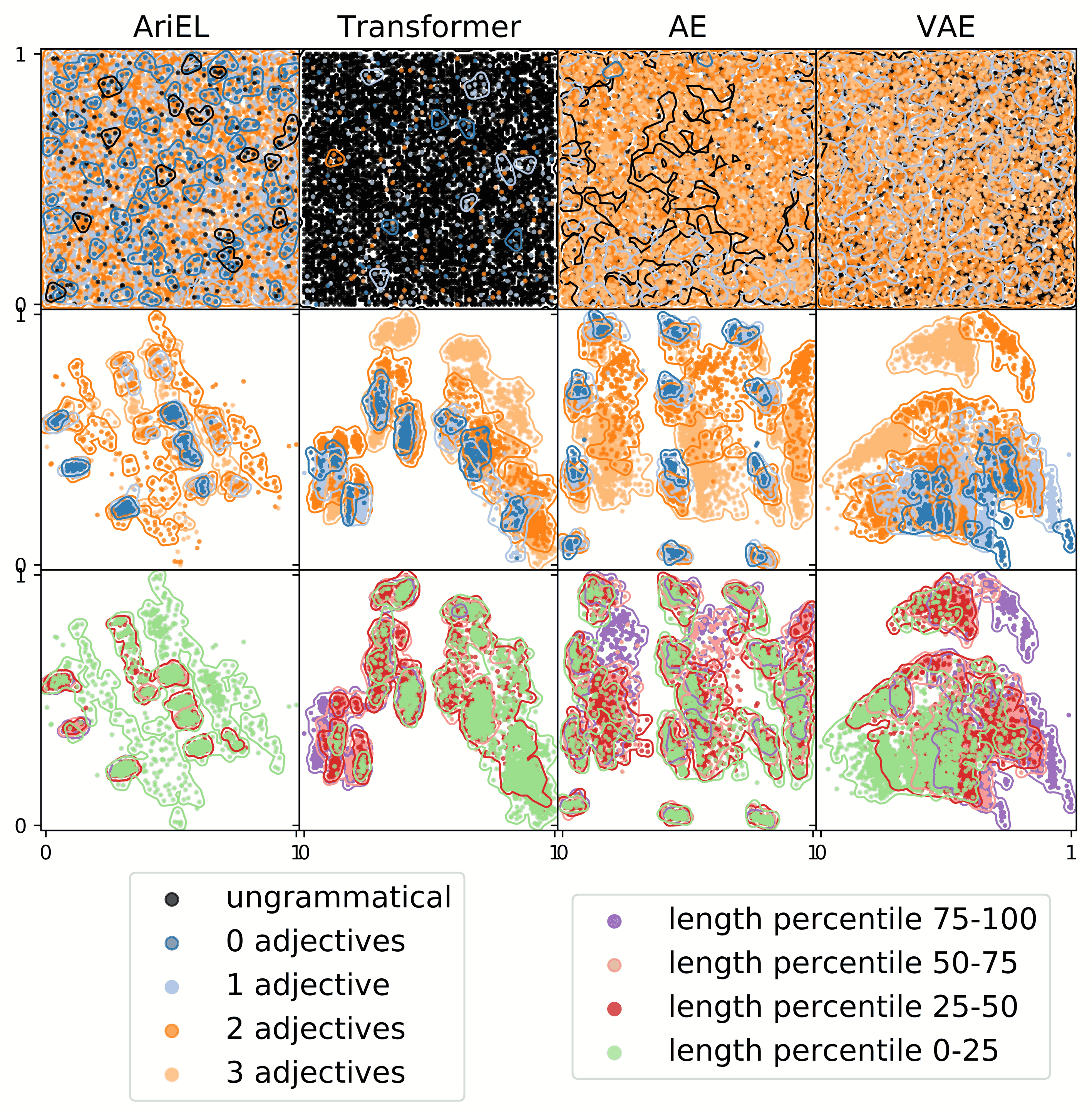}
\caption{\textbf{Random-sampling-based generation in the first row, and encoding of input sentences in the remaining rows.} A sentence is represented by a point in the latent space. First row shows the proportion of grammatically correct sentences that can be decoded by random uniform sampling the latent space. AriEL sampled almost only grammatical sentences (ungrammatical are so few that are placed on top in the plot). Transformer mainly yielded ungrammatical sentences, while AE and VAE were able to produce many grammatical sentences (ungrammatical are below, otherwise they would cover up the grammatical). Each dot is labeled according to how many adjectives the sentence generated has. Second and third rows show the clusters of points in the latent space for the test sentences as they are mapped by the encoders. All models seem to shift the clusters to some degree according to the number of adjectives in the sentence, in the second row. A similar conclusion applies to the third row, that shows where sentences of different length are encoded. For all panels, we searched subjectively for the dimensions that would better reveal some clustering, with the help of PCA. We scaled all latent representations between [0,1] for visualization.}
\label{fig:visualization45}
\end{figure}

In figure \ref{fig:visualization45}, each dot represents a sentence in the latent space. In the first row the dot in the latent space is passed as input to the decoder, while in the second and third row the dot is the output of the encoder when the biased test sentence is fed at its input.
Two random axis in $\mathbb{R}^{d}$ are chosen for the generator, first row, while two axis were chosen subjectively among the first components of a PCA for the encoder, second and third row. In every case, the values in the latent space where normalized between zero and one to ease the visualization.
Lines are used to ease the visualization of the clusters and shifts of data with their label, since the point clouds overlap and are hard to see. The curves are constructed as concave hulls of the dots based on their Delaunay triangulation, a method called alpha shapes \cite{1056714}.

We can see in figure \ref{fig:visualization45} (first row) how easy it is to find grammatical sentences when randomly sampling the latent space for each model. AriEL practically only generates grammatical sentences and AE and VAE perform reasonably well too, while Transformer fails. AriEL failures are plot on top, to remark how few they are, while AE and VAE failures are plot at the bottom, otherwise they would hide the rest given how numerous they are.
In the same figure (rows two and three) we can observe how different methods structure the input in the latent space, each with prototypical clusters and shifts. 
The Transformer presents an interesting structure of clusters whose purpose remains unclear. Interestingly, the encoding maps seem to be more organized than the decoding ones.
All the models seem to cluster or shift data belonging to different classes at the encoding, that could be taken advantage of by a learning agent placed in the latent space. However it seems hard to use the Transformer as a generator module for an agent. The good performance of AriEL is a consequence of the fact that all the latent space is utilized, and in no directions large gaps can be observed. This can be seen in the two encoding rows, where the white spaces around the cloud of dots are consequence of the rotation performed by the PCA, otherwise all the space between 0 and 1 would be utilized by AriEL.

\begin{table}
\resizebox{\columnwidth}{!}{
\begin{tabular}{l}
\doublerule
\textbf{Input Sentences} \\
is the thing this linen carpet made of tile ? \\
is it huge and teal ? \\
is the thing transparent , huge and slightly heavy ? \\
is the object antique white , tiny and closed ? \\
\textbf{AriEL} \\
is the thing this \textcolor{NavyBlue}{lime} carpet made of tile ? \\
is it huge and \textcolor{Purple}{teachable} ? \\
is the thing transparent , huge and slightly heavy ? \\
is the object antique white , tiny and closed ? \\
\textbf{Transformer}\\
is the thing this \textcolor{NavyBlue}{stretchable} carpet made of tile ? \\
is it huge and \textcolor{NavyBlue}{magenta} ?\\
is the thing transparent , huge and slightly heavy ? \\
is the object antique white , tiny and closed ? \\
\textbf{AE} \\
is the thing this \textcolor{NavyBlue}{small toilet} made of \textcolor{NavyBlue}{laminate} ?\\
is it \textcolor{NavyBlue}{this average-sized} and \textcolor{NavyBlue}{average-sized laminate} ?\\
is the thing \textcolor{NavyBlue}{very heavy} , \textcolor{NavyBlue}{heavy} and \textcolor{NavyBlue}{very} heavy ?\\
is the object \textcolor{NavyBlue}{light pink} , \textcolor{NavyBlue}{small} and \textcolor{NavyBlue}{textured} ?\\
\textbf{VAE}\\
is the thing \textcolor{NavyBlue}{a small and textured deep stone} ?\\
is it \textcolor{NavyBlue}{the light deep bedroom} ?\\
is the thing \textcolor{Purple}{textured} , \textcolor{Purple}{textured} and \textcolor{Purple}{moderately} heavy ?\\
is the \textcolor{Purple}{thing light} , \textcolor{Purple}{moderately heavy} and \textcolor{Purple}{light green} ?\\
\vspace{.01cm} \\
\doublerule
\end{tabular}
}
\caption{\textbf{Generalization: next word prediction of unbiased sentences at test time.} An unbiased sentence is given at the input, each model encodes it, and produces a reconstruction based on that code. Color means that the word was incorrectly reconstructed. Blue means that the word made the sentences comply with the bias and purple means that the incorrect reconstruction is still unbiased. Remarkably most reconstructions seem grammatically correct. AriEL keeps volumes for all sequences by construction, even if those volumes can be very small, however, in practice it made errors as well. Some of its failed reconstructions comply with the training bias, some do not, as it can be seen by the color code in the samples shown. Transformer performs remarkably well, probably given the fact that it is a word vector encoding so it is more robust to word level noise than pure sentence vector encodings. Interestingly the errors made tend to make the sentence comply to the training bias. AE produced only biased sentences, in blue, whose structure resembled the unbiased ones. VAE behaved similarly, producing less similar sentences, and more unbiased ones.}
\label{tab:generalization}
\end{table}

\begin{table}
\resizebox{\columnwidth}{!}{
\begin{tabular}{l}
\doublerule
\textbf{AriEL} \\
is the object that tiny very light set ? \\
is the thing a tiny destroyable abstraction ? \\
is the thing this mint cream textured organic structure ? \\
is the object this small large wearable textile ? \\
\textbf{Transformer}\\
\textcolor{BrickRed}{is the thing slightly heavy heavy stone squeezable} \\ \textcolor{BrickRed}{closed sea heavy ?}\\
\textcolor{BrickRed}{is it pale lime executable executable shallow decoration} \\ \textcolor{BrickRed}{drab turquoise , heavy and potang ?}\\
\textcolor{BrickRed}{is it an tomato slot box made of decoration facing stone ?}\\
\textcolor{BrickRed}{is the thing short and spring heavy slightly heavy potang ?}\\
\textbf{AE} \\
is the object that light light laminate ? \\
is the thing a light , small and small laminate ? \\
is the thing that tiny small decoration stone ? \\
is the object the average-sized , textured and \\ average-sized laminate ? \\
\textbf{VAE}\\
is the thing a light and deep office ?\\
\textcolor{BrickRed}{is it light , light and light and pink ?}\\
is the object dark , light and pink ?\\
is the object a light deep living room ?\\
\vspace{.01cm}\\
\doublerule
\end{tabular}
}
\caption{\textbf{Generation: output of the decoder when sampled uniformly in the latent space.} Red defines grammatically incorrect generations according to the CFG the models are trained on. AriEL produces an extremely varied set of grammatically correct sentences, most of which keep the bias of the training set. Transformer reveals itself to be hard to control via random sampling of the latent space, since it almost never produces correct sentences with this method. AE and VAE manage to produce several different sentences, the latter producing more non grammatical, but as well more varied grammatical ones.
}
\label{tab:generation}
\end{table}

\subsection{Quantitative Evaluations}
\label{sec:results_quantitative}

\begin{figure*}
\centering
\includegraphics[width=.95\linewidth]{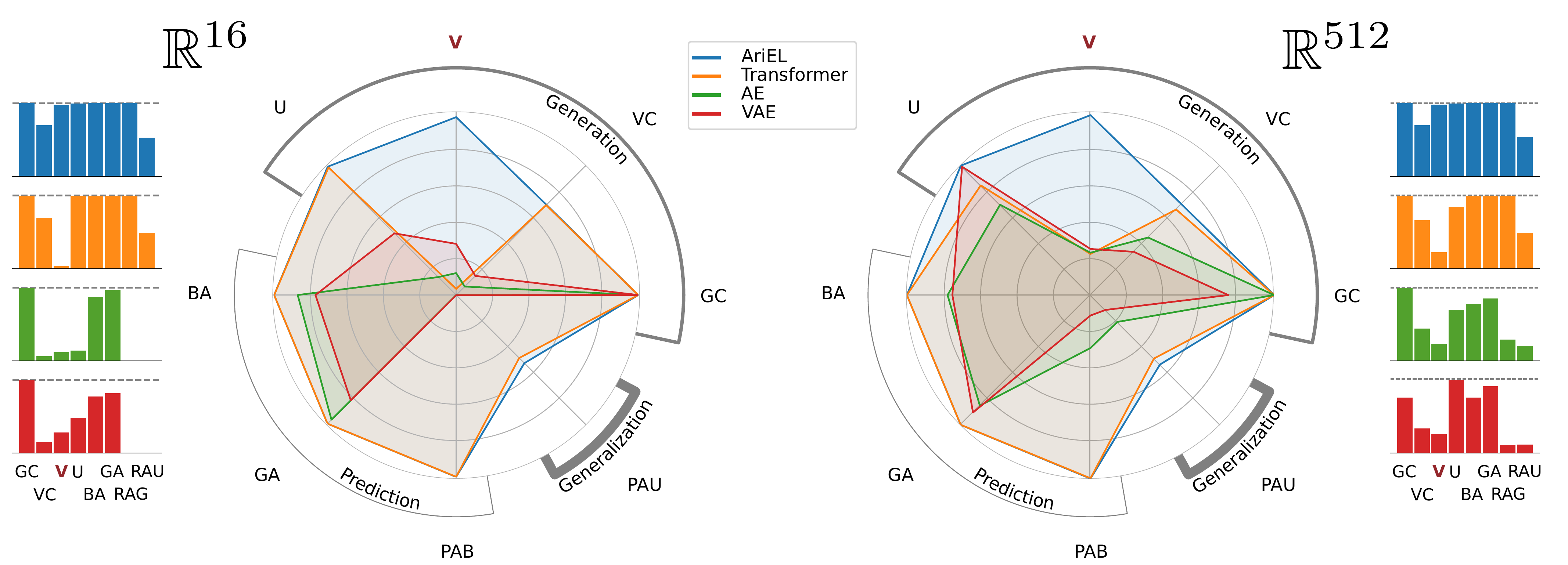}
\caption{\textbf{Radar Chart of the Quantitative Assessment. Latent space of $\mathbb{R}^{16}$ on the left and $\mathbb{R}^{512}$ on the right.} Training was performed on biased sentences. The metrics are defined in Methodology: Generalization is measured by \textit{prediction accuracy of unbiased} sentences (PAU), Prediction by \textit{prediction accuracy of biased} sentences (PAB), \textit{grammar accuracy} (GA) and \textit{bias accuracy} (BA) and Generation by \textit{uniqueness} (U), \textit{validity} (V), \textit{vocabulary coverage} (VC) and \textit{grammar coverage} (GC).
AriEL excels in all the 8 metrics. Most importantly AriEL outperforms every other method in Generation Validity (V) and it doesn't require a large latent space to do so ($\mathbb{R}^{16}$ similar to $\mathbb{R}^{512}$). VAE performs remarkably well at generating unique and grammatical sentences (validity, V) when the latent space is small ($\mathbb{R}^{16}$), probably given the volume-code nature of the method. Transformer performs exceptionally at not overfitting in the reconstruction tasks and generalizing, it manages to cover all grammar rules, even with a very small number of parameters ($\mathbb{R}^{16}$). Transformer proved to be an inefficient generator using random sampling as input (validity) but improved with a larger latent space. For a larger latent space of $\mathbb{R}^{512}$, AE and VAE overfit less (PAU and PAB) and improve their Generation (V). 
\label{fig:radars}
}

\end{figure*}

The results of the quantitative study are shown graphically in figure \ref{fig:radars} and in table \ref{tab:data}. AriEL outperforms or closely matches every other method for all the 8 measures, remarkably outperforming by a large margin every other alternative for validity, which stands for unique and grammatical sentences generated, and is the most important of the metrics for the toy dataset.
Transformer performs remarkably well at not overfitting and it is able to reconstruct biased and unbiased sentences better than the other non-AriEL methods. It does so even in the under-parameterized version (small latent space, 16-dimensions). It  manages to cover all grammar rules in generation but it performs very poorly at generating a diverse set of valid sentences from uniform random sampling. Remarkably, it only needed one iteration through the data to achieve almost perfect validation accuracy, without losing performance when we trained for the complete set of 10 epochs.
The 16-dimensional VAE despite the poor generalization to the biased test set and the unbiased test set, figure \ref{fig:radars}, results in the best non-AriEL generator, measured by validity. In this case, it could be said that the conflict between cross-entropy and KL divergence, encouraged the VAE to look for sentences that were outside the bias of the training set, since it was able to produce more grammatically correct sentences, albeit unbiased, than AE.

Increasing the learned parameters by moving from $d=16$ to $d=512$, had no effect on Transformer, that was already excellent in several of the metrics, apart from a significant improvement in validity. However, a larger latent space and the increase in number of parameters that followed, was necessary to have an AE and VAE that did not overfit (better PAU and PAB).

It is not completely fair to claim that AriEL is doing some sort of generalization however. It should rather be understood that AriEL keeps the volume for every possible sequence of symbols available, even if with a negligible size. Those volumes can easily be found and tracked having access to AriEL's Language Model. 

When trained on human sentences, on the GuessWhat?! dataset, the patterns that arise are similar and AriEL achieves again the best validity. Every approach seems to generate more unique sentences than AriEL, but the fraction of them that is a good generation is very small. Less than $6\%$ of the unique sentences generated by AE, VAE and Transformer are in the training set, while AriEL achieves $22.47\%$ and more.

\begin{figure}[!htb]
\centering
\includegraphics[width=1.\linewidth]{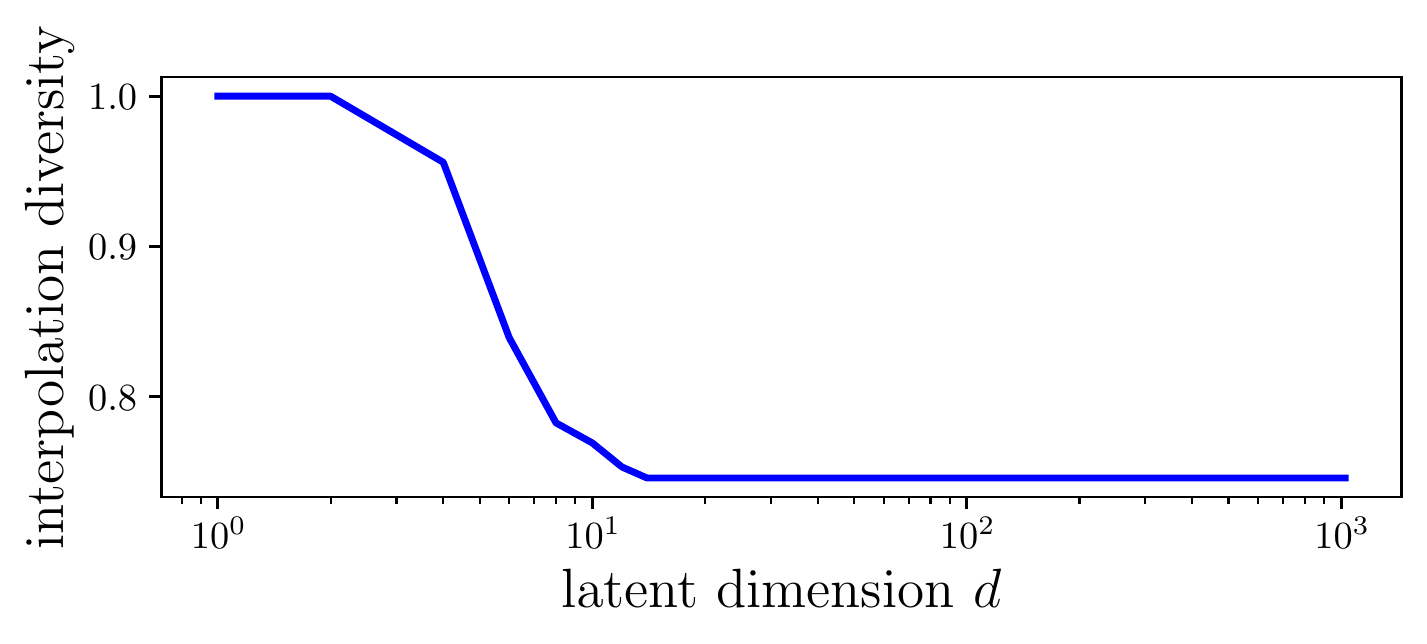}
\caption{\textbf{Interpolations between random points in the latent space of AriEL, and diversity of the sentences generated in between.} For low dimensions all sentences are very densely packed, and in the extreme of one dimension, all sentences are found following one given dimension. As the dimensionality increases, the sentences are redistributed in $ [0,1]^d$ and less sentences are found in a given direction. The lower bound at 0.746 is related to the vocabulary size and the language complexity. 
}
\label{fig:imterpolations}
\end{figure}

As  we  can  see in the interpolation diversity study in figure \ref{fig:imterpolations},  for  low $d$,  we  have  to  pass  through  many  sentences  in  between two random points in the latent space, while as we augment the dimensionality, we distribute the sentences  in  different  directions.  Therefore  we  find  less sentences  when  we  move  on  the  direction  defined  by  the straight line between two random points. The specific curve, lower threshold and speed of decay, will vary for different vocabulary sizes and given the complexity of the language learned, but the shape would be expected to remain similar.

\begin{table*}[!ht]
\resizebox{.94\width}{!}{%
\begin{tabular}{@{\extracolsep{-4pt}}rc|cccc|ccc|c@{}}
&&\multicolumn{4}{c|}{Generation} & \multicolumn{3}{c|}{Prediction} & \multicolumn{1}{c}{Generalization} \\ 
& \textbf{param} & \textbf{\shortstack{grammar \\ coverage}} & \textbf{\shortstack{vocabulary \\ coverage}} & \textbf{\shortstack{\textcolor{BrickRed}{validity}} } & \textbf{uniqueness} & \textbf{\shortstack{bias \\ accuracy}} & \textbf{\shortstack{grammar \\ accuracy} } & \textbf{\shortstack{prediction \\ accuracy \\ biased} } & \textbf{\shortstack{prediction \\ accuracy \\ unbiased} } \\
$d=16$ &&&&&&&&& \\
\textbf{AriEL} & 237K & \textbf{100.0} $\pm$ \ 0.0\% & \textbf{70.4} $\pm$ \ 0.2\% & \textbf{97.6} $\pm$ \ 0.2\% & \textbf{99.7} $\pm$ \ 0.1\% & \textbf{100.0} $\pm$ \ 0.0\% & \textbf{100.0} $\pm$ \ 0.0\% & \textbf{100.0} $\pm$ \ 0.0\% & \textbf{53.1} $\pm$ \ 0.4\% \\
\textbf{Transformer} & 258K & \textbf{100.0} $\pm$ \ 0.0\% & \textbf{70.1} $\pm$ \ 0.8\% & 4.7 $\pm$ \ 2.7\% & 99.1 $\pm$ \ 0.5\% & 99.98 $\pm$ \ 0.01\% & 99.95 $\pm$ \ 0.02\% & 99.92 $\pm$ \ 0.02\% & 49.0 $\pm$ \ 0.1\% \\
\textbf{AE} & 258K & \textbf{100.0} $\pm$ \ 0.0\% & 6.89 $\pm$ \ 0.7\% & 11.5 $\pm$ \ 4.2\% & 13.9 $\pm$ \ 5.1\% & 89.5 $\pm$ \ 2.3\% & 98.0 $\pm$ \ 1.7\% & 0.0 $\pm$ \ 0.1\% & 0.0 $\pm$ \ 0.1\% \\
\textbf{VAE} & 258K & \textbf{100.0} $\pm$ \ 0.0\% & 11.5 $\pm$ \ 2.6\% & 16.0 $\pm$ \ 9.2\% & 24.3 $\pm$ \ 14.8\% & 85.4 $\pm$ \ 5.2\% & 85.1 $\pm$ \ 8.8\% & 0.0 $\pm$ \ 0.1\% & 0.0 $\pm$ \ 0.1\% 
\vspace{.02cm} \\
$d=512$  &&&&&&&&& \\
\textbf{AriEL} & 237K & \textbf{100.0} $\pm$ \ 0.0\% & \textbf{70.2} $\pm$ \ 0.3\% & \textbf{97.9} $\pm$ \ 0.2\% & \textbf{99.8} $\pm$ \ 0.1\% & \textbf{100.0} $\pm$ \ 0.0\% & \textbf{100.0} $\pm$ \ 0.0\% & \textbf{100.0} $\pm$ \ 0.0\% & \textbf{53.2} $\pm$ \ 0.3\% \\
\textbf{Transformer} & 9M & \textbf{100.0} $\pm$ \ 0.0\% & 67.3 $\pm$ \ 0.9\% & 17.2 $\pm$ \ 6.3\% & 87.2 $\pm$ \ 7.5\% & \textbf{99.99} $\pm$ \ 0.01\% & 99.91 $\pm$ \ 0.03\% & 99.86 $\pm$ \ 0.05\% & 49.0 $\pm$ \ 0.1\% \\
\textbf{AE} & 120M & \textbf{100.0} $\pm$ \ 0.0\% & 39.3 $\pm$ \ 6.0\% & 21.0 $\pm$ \ 11.8\% & 71.8 $\pm$ \ 5.6\% & 82.2 $\pm$ \ 3.5\% & 86.8 $\pm$ \ 1.3\% & 34.7 $\pm$ \ 11.4\% & 24.4 $\pm$ \ 6.0\% \\
\textbf{VAE} & 120M & 85.0 $\pm$ \ 12.6\% & 28.9 $\pm$ \ 2.4\% & 26.5 $\pm$ \ 2.4\% & 95.2 $\pm$ \ 3.8\% & 73.8 $\pm$ \ 2.2\% & 89.5 $\pm$ \ 2.8\% & 4.3 $\pm$ \ 3.7\% & 4.9 $\pm$ \ 3.6\% \\
\vspace{-.2cm} \\
\end{tabular}%
}
\vspace{.5cm}
\caption{\textbf{Evaluation of continuous sentence embeddings.} Complete results for the different methods and the different proposed measures, for varying dimensionality $d$ of the latent space, $d = 16$ and $d= 512$. Each experiment is run 5 times, we report the mean and the variance for each configuration. We comment on the content of this table in section \ref{sec:results_quantitative} and they are plot in figure \ref{fig:radars}. Our proposed method, AriEL, achieves almost perfect performance in almost all the metrics that we defined, especially in Generation validity, which quantifies how many random samples gave a unique and grammatical sentence at the output of the decoder. Transformer performed exceptionally well even in the under-parameterized case, with a 16-dimensional latent space, except for validity. All methods improved their performance with a larger latent space, and when over-parameterized ($d= 512$), particularly in validity, but still achieved less than one third the performance of AriEL. VAE is consistently the second best performer in validity, supporting our hypothesis, that volume coding facilitates retrieval of information by random sampling.
}
\label{tab:data}
\end{table*}

\begin{table}[H]
\begin{tabular}{rc|ccc}
& \textbf{param} & \textbf{\shortstack{prediction \\ accuracy}} & \textbf{uniqueness} & \textbf{\textcolor{BrickRed}{validity}} \\
&&&& \\
$d=16$ &&&&\\
\textbf{AriEL} & 2,901K & 92.56\% & 57.41\% & \textbf{29.59\%} \\
\textbf{Transformer} & 588K & 87.91\% & \textbf{96.75\%} & 1.96\%  \\
\textbf{AE} & 2,787K & 11.97\% & 13.68\% & 2.67\%    \\
\textbf{VAE} & 2,787K & 13.15\% & 6.26\% & 1.61\%    \\
&&&& \\
$d=512$  &&&&\\
\textbf{AriEL} & 2,901K & 92.54\% & 55.46\% & \textbf{22.47\%} \\
\textbf{Transformer} & 18,809K & 86.9\% & 50.33\% & 5.94\% \\
\textbf{AE} & 4,900K & 15.45\% & 12.4\% & 2.68\% \\
\textbf{VAE} & 5,425K & 32.51\% & \textbf{98.51}\% & 0.2\%
\end{tabular}
\vspace{.5cm}
\caption{\textbf{Performance on the GuessWhat?! Questioner data.} For the real dataset the pattern is repeated. AriEL is the approach that generates the more diverse set of unique sentences from the training set. The underparameterized Transformer 16 gave much better results than when its hyper-parameter $n_{parallel \ layers}$ was increased from 2 to 20 to increase its learnable parameters from 588K to 2,666K.}
\label{tab:dataGW}
\end{table}
\section{Discussion}
\label{sec:discussion}

\noindent\textbf{Brief summary of the results.} Transformer has proven to be exceptional at not to overfit during training, with a very quick learning. It did so requiring significantly less parameters than classical approaches (AE and VAE).
Embeddings learned by the Transformer encoding revealed interesting structures that remain unexplained (figure \ref{fig:visualization45}). Those structures are projected as key and value to the multihead attention \citep{vaswani2017attention}.
However it is hard to find the language that it learned by randomly sampling the latent space. We hypothesize that this is due to its word vector embedding nature: even if its latent dimension was $16$ on the word level, at the sentence level it was $20\cdot16 = 320$ given that an artificial padding to the maximum sentence length ($25$) was introduced to be able to perform several of the quantitative studies. A consequence of very high dimensional spaces is that points tend to be all equally distant to each other. The explanation might therefore be that ungrammatical sentences outnumber grammatical ones, and therefore were easier to find.
AE and VAE needed a high dimensional latent space and therefore a larger number of parameters to be able to generalize to the biased and unbiased test sets (PAU and PAB). However they proved to be decent generators for a small latent space.

\noindent\textbf{AriEL latent space $d$ is a free parameter.} A detail that is worth to stress, is that the size $d$ of the latent space of AriEL can be defined at any time, for a fixed Language Model. It could therefore be conceivable to control it with a learnable parameter, with the activity of another neuron, or as another function of the input. In fact, as we augment the dimension, the volumes will tend to have more neighbouring volumes that represent different sentences as confirmed by the interpolation study, figure \ref{fig:imterpolations}.
It could have implications as well during training to have a gradient that can rely more on its angle than on its magnitude.

\noindent\textbf{What to choose for a learning agent with a language module?} In the case of a learning agent that needs a language model to interact with other agents, our study suggests that it will benefit from AriEL to generate a diverse language (generation). Interestingly it outperformed the other methods in the toy dataset but as well on the real data.
In order to encode language, it might benefit from any of the methods, or an ensemble of them.
As well, in this work we trained AriEL's Language Model (LM) before utilizing it inside AriEL. If AriEL was used as a module of a larger architecture, it would be necessary to design a proper pseudo-gradient to train end to end its LM. This can be the case for the training of a GAN for text \cite{goodfellow2014generative, fedus2018maskgan}, that typically fails due to (1) mode collapse, (2) given the non differentiability of argmax. AriEL would make sure the availability of a wide and complex language model, decreasing the chances of mode collapse, and given that the next token selection is not necessarily done through argmax.

\noindent\textbf{Partial evidence for volume codes.} As a consequence of the experiments performed, it is our impression that the volume aspect of AriEL is to be held responsible of its success, and that's why we provide it as a definition that could reveal itself more useful than the algorithm we present. We have provided some evidence on how volume coding can be beneficial for retrieval of stored information that is composed of discrete symbols, and variable length, by means of random sampling, in contrast with simply distributed representations. It is in fact AriEL to be the method that generates more valid sentences, a method that performs explicit volume coding. VAE is the second when trained over the toy dataset, a method that performs implicit volume coding, encouraged by the loss. However it performed rather poorly on the real dataset. Point encodings can still provide an advantage for example in classification. In our case, a point encoding would be represented by Transformer, and the classification task the PAU and PAB metrics: it trained fast, without overfitting, and underparametrized, but the absence of dense volumes made it hard to retrieve the stored information using uniform random samples.

\section{Conclusion and Future Work}


We proposed AriEL, a volume mapping of language into a continuous hypercube. It provides a latent organization of language that excels at several important metrics related to the use of language, giving special emphasis to being able to generate many unique and grammatically correct sentences sampling uniformly the latent space, what we call \textit{valid} sentences. AriEL fuses together arithmetic coding and k-d trees to construct volumes that preserve the statistics of a dataset. In this way we construct a latent representation that explicitly assigns a data sample to a volume, instead of a point. When compared to standard techniques it performs favorably in generation, prediction and generalization.

Moreover, we used a manually designed context-free grammar (CFG) to generate our own large-scale dataset of sentences, and we assigned a random bias using a randomly generated adjacency matrix of words that can appear together. This allows us to (1) automatically check if the language generated by the models belongs to the grammar used and (2) understand if different deep learning methods can grasp notions of grammar separately from other notions of bias. We used a dataset of real human interactions to make sure the findings would hold for a larger vocabulary, and a less strict grammar.

Recurrent-based continuous sentence embeddings largely overfit the training data and only cover a small subset of the possible language space, particularly when the size of the latent space is small. They also fail to learn the underlying CFG and generalize to unbiased sentences from that CFG. However they manage to generate quite a few diverse valid sentences. Transformer managed to avoid overfitting even after being overtrained, proving its robustness.
It performed a remarkable generalization to the unbiased data. However it proves hard to use as a generator from the continuous latent space using random sampling.

We stress that volume based codes can provide an advantage over point codes in generation tasks, or sampling tasks to call it differently.
Moreover, our method allows us to sample/generate in theory the same probability distribution as the training set and in practice a much more diverse set of sentences, as demonstrated on the toy dataset and on the human dataset. 

Our planned next step is to use AriEL as a module in a learning agent. It would be interesting to apply AriEL to k-Nearest Neighbour, and optimize it for very high dimensional latent spaces. This study has been performed for dialogue based language generation, which implies short sentences. It would be useful for the NLP community to understand if this method generalizes to the compression of longer texts.

\subsubsection*{Acknowledgments}
The authors would like to thank the ERA-NET (CHIST-ERA) and FRQNT for funding this research as part of the IGLU project. NVIDIA Corporation supported this research with the donation of a Titan X and Tesla K40. We wish to thank  Etienne Richan, Eric Plourde, Jacob Lavoie, Maryam Hosseini, Ahmad El Ferdaoussi, Louis Comm\`ere, Emmanuel Calvet, Mathilde Brousmiche, Ismaël Balafrej and Matin Azadmanesh for useful comments and proof reading.

\clearpage


\bibliographystyle{IEEEtran}
\bibliography{references}



\clearpage
\beginsupplement

\section*{Supplementary Material}

\section{Context-free grammar (CFG) used in the experiments}
\label{app:cfg}

The context free grammar used to generate the biased and unbiased sentences is composed by the following rules:

\begin{minipage}{\linewidth}
\scriptsize
\begin{lstlisting}
s -> q

q -> qword adjective ',' adjective 'and' adjective '?'
q -> qword adjective 'and' adjective '?'
q -> qword adjective '?'
q -> qword 'made' 'of' noun_material '?'
q -> qword preposition np '?'
q -> qword np '?'

np -> determiner adjective adjective adjective noun
np -> determiner adjective ',' adjective 'and' adjective noun
np -> determiner adjective 'and' adjective noun 'made' 'of' noun_material
np -> determiner adjective adjective noun
np -> determiner adjective 'and' adjective noun
np -> determiner adjective noun 'made' 'of' noun_material
np -> determiner noun 'made' 'of' noun_material
np -> determiner adjective noun
np -> determiner noun

qword -> 'is' 'it' | 'is' 'the' 'object' | 'is' 'the' 'thing'
noun -> noun_object | noun_material | noun_roomtype
preposition -> preposition_material

adjective -> adjective_color | adjective_affordance | adjective_overall_size | 
             adjective_relative_size | adjective_relative_per_dimension_size | 
             adjective_mass | adjective_state | adjective_other

noun_object -> 'accordion' | 'acoustic' 'gramophone' | 'bar' | 'barrier' |
               'basket' | 'outdoor' 'lamp' | 'outdoor' 'seating' | ...

noun_material -> 'bricks' | 'carpet' | 'decoration' 'stone' | 'facing' 'stone' | 
                 'grass' | 'ground' | 'laminate' | 'leather' | 'wood' | ...

noun_roomtype -> 'aeration' | 'balcony' | 'bathroom' | 'bedroom' | 'boiler' 'room' | 
                 'garage' | 'guest' 'room' | 'hall' | 'hallway' | 'kitchen' | ...

determiner -> 'a' | 'an' | 'that' | 'the' | 'this'

preposition_material -> 'made' 'of'

adjective_color -> 'antique' 'white' | 'magenta' | 'maroon' | 
                   'slate' 'gray' | 'white' | 'yellow' | ...
                   
adjective_affordance -> 'actable' | 'addable' | 'addressable' | 'deliverable' | 
                        'destroyable' | 'dividable' | 'movable' | ...
                        
adjective_size -> adjective_overall_size | adjective_relative_size |
                  adjective_relative_per_dimension_size
                  
adjective_overall_size -> 'average-sized' | 'huge' | 'large' | 'small' | 'tiny'
adjective_relative_size -> 'average-sized' | 'huge' | 'large' | 'small' | 'tiny'
adjective_relative_per_dimension_size -> 'deep' | 'narrow' | 'shallow' | 
                                         'short' | 'tall' | 'wide'

adjective_mass -> 'heavy' | 'light' | 'moderately' 'heavy' | 'moderately' 'light' | 
                  'slightly' 'heavy' | 'very' 'heavy' | 'very' 'light'
adjective_state -> 'closed' | 'opened'
adjective_other -> 'textured' | 'transparent'
\end{lstlisting}
\end{minipage}

\clearpage

\section{Size of the language space}

From the CFG used in the experiment, it is possible to extract a total of 15,396 distinct grammar rules, some are shown below. However, for simplicity, we defined only 4, related to the number of adjectives in it. In the case of the unbiased dataset, those rules can produce a total of 9.81e+18 unique sentences. The total number of unique sentences for the biased dataset is expected to be an order of magnitude smaller.

\begin{minipage}{\linewidth}
\scriptsize
\begin{lstlisting}
[qword, prep_material, determiner, adj_state, 'and', adj_other, noun_roomtype, '?']
[qword, prep_spatial, determiner, adj_other, adj_state, adj_state, noun_object, '?']
[qword, determiner, adj_other, ',', adj_mass, 'and', adj_affordance, noun_roomtype, '?']
[qword, determiner, adj_relative_per_dimension_size, adj_overall_size, noun_object, '?']
[qword, determiner, adj_overall_size, ',', adj_state, 'and', adj_state, noun_material, '?']
[qword, prep_spatial, determiner, adj_other, adj_mass, adj_affordance, noun_material, '?']
[qword, adj_state, 'and', adj_relative_size, '?']
[qword, prep_material, determiner, adj_mass, adj_other, adj_other, noun_material, '?']
[qword, prep_spatial, determiner, adj_state, adj_other, adj_color, noun_object, '?']
[qword, determiner, adj_relative_size, 'and', adj_overall_size, noun_material, '?']
[qword, determiner, adj_state, adj_overall_size, adj_other, noun_roomtype, '?']
[qword, determiner, adj_other, adj_state, adj_mass, noun_material, '?']
[qword, determiner, adj_overall_size, 'and', adj_other, noun_material, '?']
[qword, determiner, adj_color, adj_other, noun_object, '?']
[qword, prep_spatial_rel, determiner, adj_mass, adj_color, noun_roomtype, '?']
[qword, determiner, adj_state, 'and', adj_relative_size, noun_object, '?']
[qword, determiner, adj_color, adj_color, adj_relative_size, noun_material, '?']
[qword, determiner, adj_affordance, noun_object, '?']
[qword, determiner, adj_other, adj_other, adj_state, noun_roomtype, '?']
\end{lstlisting}
\end{minipage}

\section{Example of sentences generated from the CFG}

\subsection{Biased sample sentences}

\label{app:grounded}

\begin{itemize}
\item is it large , light yellow and light ?
\item is it white , deep pink and average-sized ?
\item is it a light , huge and shallow laminate ?
\item is the object average-sized and light ?
\item is the object fashionable , ghost white and pale turquoise ?
\item is the thing huge , huge and khaki ?
\item is the thing small , ignitable and very light ?
\item is the object a notable very light orange carpet ?
\item is the object this small wood made of facing stone ?
\item is the object a textured and combinable floor cover made of laminate ?
\end{itemize}

\subsection{Unbiased sample sentences}

\begin{itemize}
\item is the object the huge tiny lovable guest room ?
\item is the object the closed closed transparent textile ?
\item is the thing a transparent , narrow and slightly heavy textile ?
\item is it steerable , dark orange and light ?
\item is it gray , very heavy and textured ?
\item is it closed , heavy and moderately light ?
\item is it transparent , transformable and moderately light ?
\item is the thing average-sized and dark red ?
\item is the thing large and deep garage ?
\item is it that slightly heavy stucco made of grass ?
\end{itemize}

\clearpage
\section{Vocabulary}

\begin{table}[H]
\begin{tabular}{|l|r|l|}
\hline
\multicolumn{1}{|c|}{\textbf{Annotation}} & \multicolumn{1}{|c|}{\textbf{Nb. of classes}} & \multicolumn{1}{|c|}{\textbf{Example of classes}} \\ \hline
Noun & 86 & air conditioner, mirror, window, door, piano \\ \hline
WordNet category \citep{Miller1995}& 580 & instrument, living thing, furniture, decoration \\ \hline
Location & 24 & kitchen, bedroom, bathroom, office, hallway, garage \\ \hline
Color & 139 & red, royal blue, dark gray, sea shell \\ \hline
Color property & 2 & transparent, textured \\ \hline
Material & 15 & wood, textile, leather, carpet, decoration stone \\ \hline
Overall mass & 7 & light, moderately light, heavy, very heavy \\ \hline
Overall size & 4 & tiny, small, large, huge \\ \hline
Category-relative size & 10 & tiny, small, large, huge, short, shallow, narrow, wide \\ \hline
State & 2 & opened, closed \\ \hline
Acoustical capability & 3 & sound, speech, music \\ \hline
Affordance & 100 & attach, bend, divide, play, shake, stretch, wear \\ \hline
\end{tabular} \vspace{1ex}
\caption{Description of vocabulary used.}
\label{tb:annotations}
\end{table}

\end{document}